\documentclass{article} % For LaTeX2e
\usepackage{iclr2027_conference,times}

\usepackage{amsmath,amsfonts,bm}

\def\eqref#1{equation~\ref{#1}}
\def\1{\bm{1}}

\DeclareMathAlphabet{\mathsfit}{\encodingdefault}{\sfdefault}{m}{sl}
\SetMathAlphabet{\mathsfit}{bold}{\encodingdefault}{\sfdefault}{bx}{n}

\usepackage{hyperref}
\usepackage{url}
\usepackage{multirow}
\usepackage{float}
\usepackage{booktabs}
\usepackage{graphicx}
\title{Holi-Spatial 360: Generative Reconstruction from Sparse Panoramas for Spatial Rendering and Intelligence}

\author{%
\bfseries
Yuanyuan Gao$^{1,*}$, Zhirui Liu$^{2,*}$, Yuning Gong$^{3}$, Yifei Liu$^{4}$, \\
\bfseries
Yifan Zhan$^{5}$, Kuixuan Jiao$^{6}$, Xiaolong Zhou$^{4}$, Yifan Yang$^{7}$, \\
\bfseries
Hongjie Zhang$^{4}$, Xue Yang$^{7}$, Dan Xu$^{1}$, and Zhihang Zhong$^{7,\dagger}$ \\
\normalfont
$^{1}$The Hong Kong University of Science and Technology, $^{2}$ShanghaiTech University, \\
$^{3}$Tencent, $^{4}$Shanghai Artificial Intelligence Laboratory, $^{5}$The University of Tokyo, \\
$^{6}$Northwestern Polytechnical University, $^{7}$Shanghai Jiao Tong University \\
\vspace{-1em}
\href{https://visionary-laboratory.github.io/Holi-spatial-360/}
{\textbf{Project Page:} visionary-laboratory.github.io/Holi-spatial-360/}
}

\usepackage{booktabs}
\usepackage{makecell}
\usepackage{siunitx}
\usepackage{pifont}

\newcommand{\cmark}{\ding{51}}
\newcommand{\xmark}{\ding{55}}

\iclrfinalcopy % Uncomment for camera-ready version, but NOT for submission.
\begin{document}

\maketitle

% Remove ICLR publication status for arXiv preprint
\lhead{}
% Author contribution footnote
\begingroup
\makeatletter
\renewcommand{\thefootnote}{}%
\long\def\@makefntext#1{\noindent#1}%
\footnotetext{%
    $^{*}$Equal contribution. \quad
    $^{\dagger}$Corresponding author.%
}
\makeatother
\endgroup

\begin{figure}[H]
    \centering
    \includegraphics[width=\linewidth]{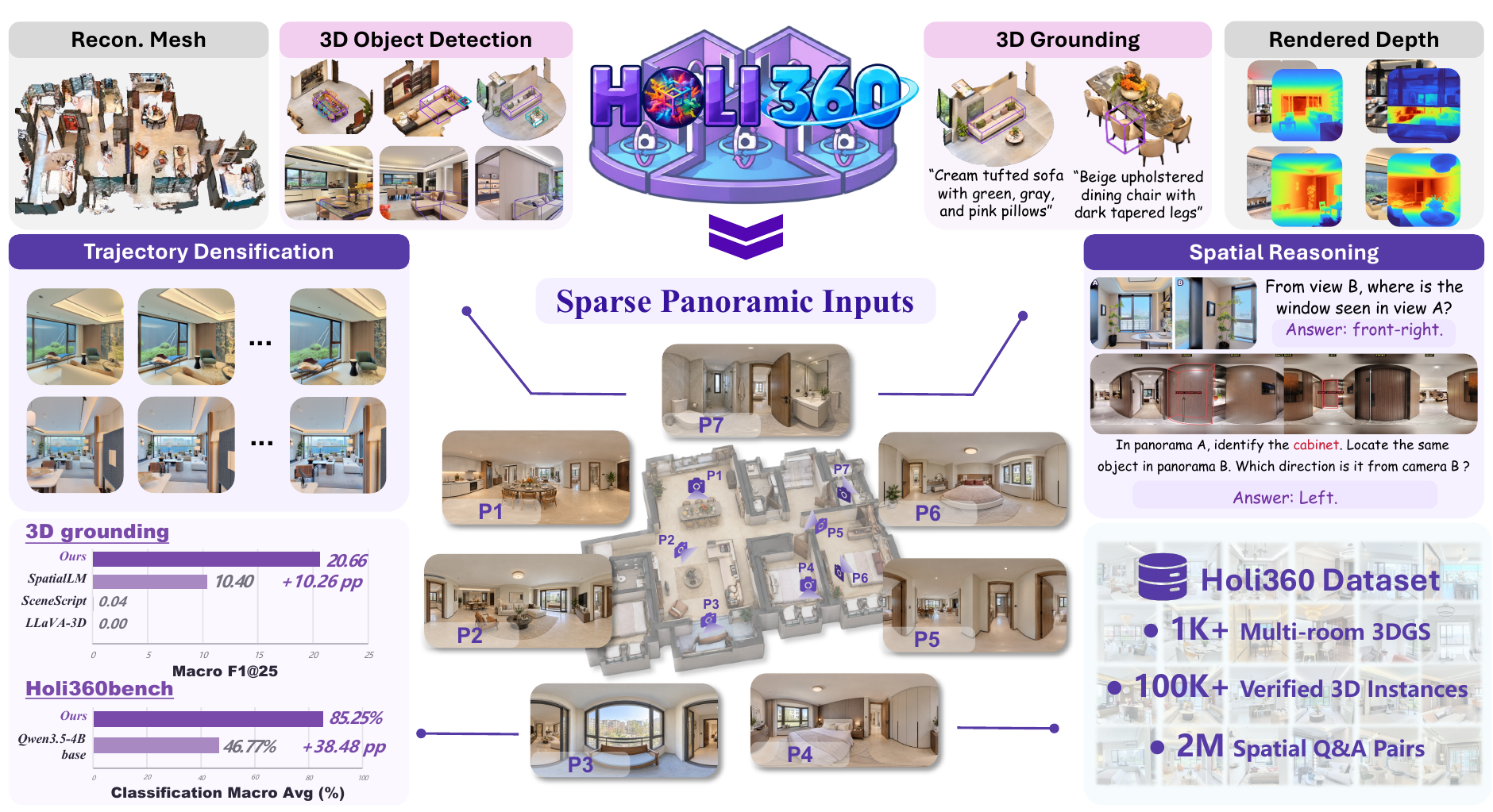}
    \caption{
        \textbf{Overview of Holi-Spatial 360.} Sparse panoramas are reconstructed into geometrically multi-room 3DGS, densified with image and video generative priors, and converted into RGB-D, 3D grounding, and spatial-reasoning supervision.
     }
    \label{fig:teaser}
\end{figure}

\begin{abstract}Spatial intelligence requires scalable 3D environments that are both photorealistic and semantically grounded, yet existing indoor datasets are dominated by room-scale captures and costly LiDAR acquisition. Online real-estate platforms offer abundant panoramas of diverse multi-room homes, but sparse viewpoints and texture-poor regions make reconstruction severely under-constrained. We introduce Holi-Spatial 360, a generative reconstruction and data curation framework for converting these panoramas into annotated multi-room 3D environments. Because reconstruction quality bounds subsequent generation, our Correspondence-Guided Geometric Optimization converts feed-forward reconstruction priors into reliable cross-view constraints and builds a coherent 3D Gaussian Splatting representation as 3D spatial memory. Adapted single-image and video priors then restore appearance and complete continuous trajectories while preserving observed content. The reconstructed scenes support arbitrary RGB-D rendering and an automated curation pipeline that combines end-to-end 3D grounding with VLM-based verification. Holi360 contains over 1K multi-room 3DGS scenes, more than 100K instance annotations, and 2M spatial question-answer pairs. We also introduce the Holi360 benchmark to evaluate the spatial understanding of VLMs in real-world, multi-room indoor environments. Experiments demonstrate superior reconstruction and curation quality under sparse panoramic inputs, while fine-tuning VLMs on Holi360 substantially improves spatial reasoning capabilities.
\end{abstract}

\section{Introduction}

Spatial intelligence aims to equip large models~\citep{Qwen3-VL,team2023gemini,Qwen2.5-VL} with the ability to perceive, ground, reason, and act in the 3D world. Its progress depends on scalable 3D environments that are both photorealistic and semantically grounded. Multi-room homes are particularly important for embodied applications, where understanding must extend beyond individual objects to layouts and relationships across interconnected spaces.

However, existing indoor datasets remain difficult to scale. ScanNet~\citep{dai2017scannet} and ScanNet++~\citep{yeshwanth2023scannet++} are dominated by room-scale captures, while building-scale datasets such as Matterport3D~\citep{Matterport3D} and HM3D~\citep{ramakrishnan2021habitat} rely on dedicated capture and processing. SAGE-3D~\citep{miao2026towards} provides semantically annotated 3DGS environments but builds upon designed 3D assets and expert annotation. Holi-Spatial~\citep{gao2026holi} automates spatial data curation from videos, yet relies on densely captured trajectories with continuous camera motion and sufficient view overlap. Diverse, real-world multi-room environments with both high-quality reconstruction and semantic supervision therefore remain costly to obtain.

Online real-estate platforms offer a largely untapped alternative: abundant 360-degree panoramas of diverse multi-room homes. Yet these panoramas are captured at only a few widely separated locations. Limited overlap, texture-poor surfaces, and extensive unobserved regions make cross-view matching and geometric reconstruction unreliable~\citep{ren2025panosplatt3r,jin2026planargs}. Recent image- and video-based generative methods can repair artifacts in novel views~\citep{wu2025difix3d+,yin2025gsfixer,de2026artifixer,wang2026artifactworld}, but they do not by themselves establish reliable geometry from sparse, multi-room panoramas. When the initial reconstruction is misaligned, generated views may drift from the observed scene and reinforce errors when distilled back into 3D.

To address these challenges, we introduce \textbf{Holi-Spatial 360}, a generative reconstruction and data curation framework that transforms sparse panoramas into semantically annotated multi-room 3D environments. Our key insight is that reconstruction quality sets the ceiling for generation. \textbf{Correspondence-Guided Geometric Optimization} uses priors from feed-forward large reconstruction models~\citep{wang2025vggt,lin2025depth} to establish cross-view constraints and optimize a coherent initial 3D Gaussian Splatting (3DGS) representation. This representation serves as persistent spatial memory for \textbf{generative trajectory densification}: image and video priors restore and complete novel views, which are distilled back into 3D while remaining grounded in the observed panoramas. The scenes support RGB-D rendering and \textbf{automated spatial data curation}, including verified 3D instances, grounding annotations, and spatial question-answer pairs. The pipeline requires only RGB panoramas, without depth capture or human-in-the-loop labeling.

From publicly available real-estate and rental panoramas, we construct \textbf{Holi360}: over 1K multi-room 3DGS scenes, dense RGB-D renderings and meshes, more than 100K verified 3D instances, and 2M spatial question-answer pairs. Reconstruction and annotation use only RGB panoramas; RealSee3D~\citep{Li2025realsee3d_data} LiDAR and semantic maps are reserved for evaluation. Experiments show improved rendering, geometry, and grounding, while VLM fine-tuning on Holi360 boosts spatial reasoning.

\section{related work}

\subsection{Scalable Indoor Environments for Spatial Intelligence}

Recent datasets such as SenseNova-SI-800K~\citep{cai2025scaling} and VST-P~\citep{yang2025visual} scale spatial supervision, but annotation count does not imply environment diversity. EmbodiedScan~\citep{embodiedscan} and SceneSplat-7K~\citep{li2025scenesplat} reuse existing scans, while HM3DSEM~\citep{yadav2023habitatmatterport3dsemanticsdataset}, HM3D-OVON~\citep{Yokoyama2024HM3DOVONAD}, and GOAT-Bench~\citep{Khanna2024GOATBenchAB} derive further annotations or tasks from HM3D~\citep{ramakrishnan2021habitat}. These efforts enrich supervision over a largely fixed pool of 3D assets. We instead scale the pool itself by reconstructing diverse, real-world multi-room homes from online panoramas.

\subsection{Generative Reconstruction from Sparse Views}

Generative priors have become a powerful remedy for sparse-view reconstruction. ReconFusion~\citep{Wu2023ReconFusion3R} regularizes NeRF optimization with image diffusion, while 3DGS-Enhancer~\citep{Liu20243DGSEnhancerEU}, Difix3D+~\citep{wu2025difix3d+}, and GenFusion~\citep{Wu2025GenFusionCT} distill restored novel views into 3D representations. Video-based methods further improve temporal consistency through scene- or reference-grounded guidance~\citep{Zhong2025TamingVD,yin2025gsfixer}, long autoregressive generation~\citep{de2026artifixer}, or large-scale restoration training~\citep{wang2026artifactworld}. Although these methods refine the resulting 3D representation, their main advances center on the generator, leaving the initial reconstruction largely unchanged. Beyond static scenes, NeoVerse~\citep{Yang2026NeoVerseE4} and One4D~\citep{Mi2025One4DU4} extend coupled generation and reconstruction to dynamic 4D scenes from monocular videos or sparse frames. PanoWorld~\citep{Jia2026PanoWorldAG} similarly maintains a dynamic 3DGS cache, but relies on single-image diffusion to update it, making independently generated views susceptible to cross-view inconsistencies similar to those observed in DiFix. In contrast, we globally optimize a unified 3DGS from observed panoramas and use generation to densify missing evidence while preserving scene identity along continuous trajectories.

\section{Method}
\begin{figure*}[t]
    \centering
    \includegraphics[width=\textwidth]{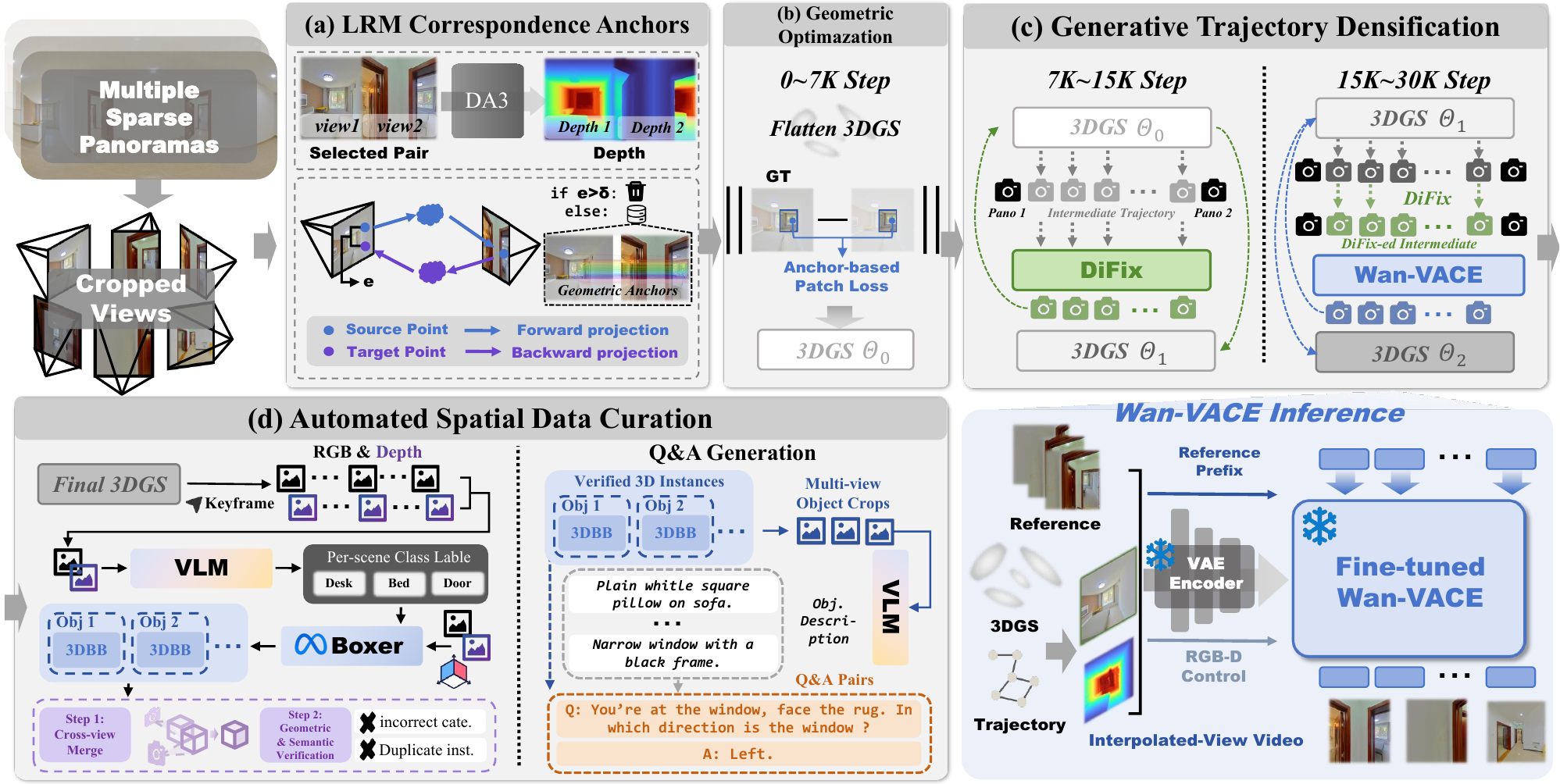}
    \caption{
    \textbf{Framework overview.} DA3 correspondences anchor 3DGS optimization; adapted DiFix and Wan-VACE~\citep{jiang2025vace} densify trajectories;
    verified RGB-D groundings yield 3D instances, captions, and spatial QA.
    }
    \label{fig:overview}
\end{figure*}

\subsection{Correspondence-Guided Geometric Optimization}
\label{sec:holi_reconstruction}

Sparse panoramic inputs provide limited overlap and often contain large
textureless regions, where conventional feature matching yields too few
reliable constraints for 3DGS optimization. We therefore leverage a frozen
feed-forward large reconstruction model (LRM) as an offline correspondence
teacher. As shown in Appendix Tab.~\ref{tab:matching_points}, conventional feature-based matchers such as i3dgs~\citep{2026immediate3DGS} and COLMAP~\citep{pan2024glomap} are insensitive to the number of input views because they operate pairwise, but produce too few reliable correspondences on sparsely textured indoor surfaces. LRMs provide substantially denser matches, although the number of forward--backward-consistent correspondences decreases when additional context views are jointly processed. We therefore select nearby view pairs with sufficient co-visibility and process each pair independently with DA3~\citep{lin2025depth}, yielding a dense set of reliable correspondence anchors for subsequent 3DGS optimization.

Given a candidate pair $(I_i,I_j)$, the LRM predicts dense geometry, relative
camera transformation $\widetilde{\mathbf{T}}_{j\leftarrow i}$, and per-pixel
confidence. For each source point $\mathbf{p}$, we obtain its target coordinate
and forward--backward error as
\begin{equation}
\begin{aligned}
\widetilde{\mathbf{q}}_{i\rightarrow j}
&=\pi_j\!\left(
\widetilde{\mathbf{T}}_{j\leftarrow i}
\pi_i^{-1}(\mathbf{p},\widetilde{D}_i(\mathbf{p}))
\right),\\
e_{ij}^{\mathrm{fb}}(\mathbf{p})
&=d_{\pi}\!\left(
\mathbf{p},
\widetilde{\mathbf{p}}_{i\rightarrow j\rightarrow i}
\right).
\end{aligned}
\label{eq:lrm_correspondence}
\end{equation}
We retain only valid, confident correspondences satisfying
$e_{ij}^{\mathrm{fb}}<\tau_{\mathrm{fb}}$ and collect them as fixed anchors
$\mathcal{M}_{ij}=\{(\mathbf{p},\mathbf{q},w_{ij})\}$, where $w_{ij}$ is determined by the LRM confidence and forward--backward error. These correspondences are computed once before training and held fixed during the initial 3DGS optimization. Similar to the multi-view geometric
regularization used in GaussianPro~\citep{Cheng2024GaussianPro3G} and
PGSR~\citep{chen2024pgsr}, we use them as fixed cross-view references
for geometric supervision. Because the correspondences are not
re-estimated from the evolving 3DGS, they help stabilize optimization
in texture-poor regions.

\subsection{In-the-Wild Generative Prior Adaptation}

\begin{figure*}[t]
    \centering
    \includegraphics[width=\textwidth]{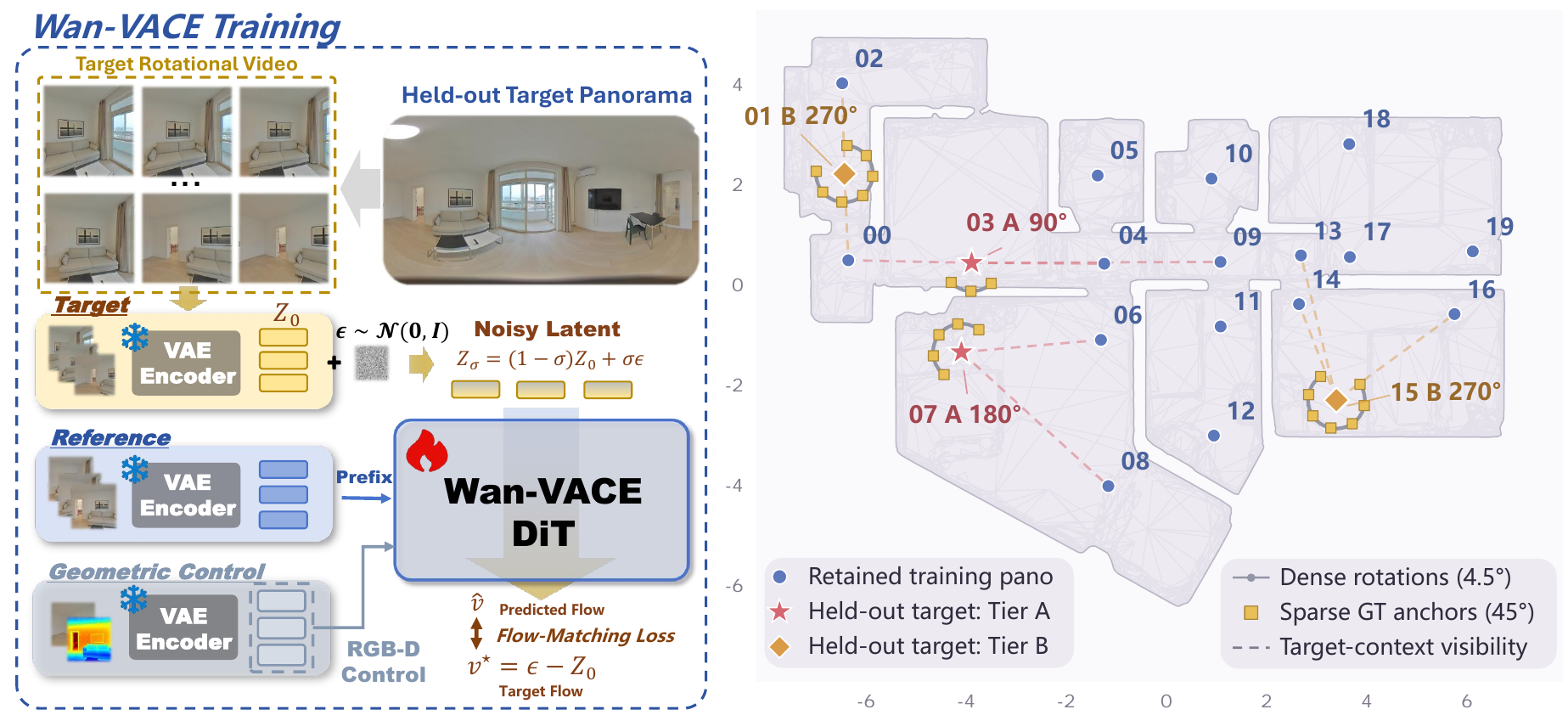}
    \caption{
    \textbf{Panorama-derived supervision.} \textbf{Left:} Held-out panoramas provide image pairs and fixed-center rotation sequences for adapting DiFix and Wan-VACE; \textbf{Right:} An example shows the train--test split and co-visibility graph.
    }
    \label{fig:data_construction}
\end{figure*}

\noindent\textbf{Training data construction.}
We construct a generative adaptation corpus from over 1K multi-room real-estate scenes. For each scene, we build a co-visibility graph over its panoramas and partition them into a connected reconstruction set
and a held-out target set. We optimize a 3DGS using only the retained panoramas, following the same reconstruction pipeline and hyperparameters
used at inference. All perspective views derived from held-out panoramas are excluded from this reconstruction. Renderings at held-out camera poses provide naturally degraded inputs, while the corresponding panoramic reprojections supply ground-truth RGB supervision. Reference views are selected from nearby retained panoramas. This construction preserves the natural artifacts caused by sparse
coverage, texture-poor regions, and viewpoint extrapolation. The right panel visualizes the split and co-visibility graph; tiers A and B use strict and permissive coverage criteria, respectively.

\noindent\textbf{Single-image restoration.}
For each sparse ground-truth anchor view $t$ sampled from a held-out panorama, we pair its 3DGS rendering $I_t^{\mathrm{render}}$ with the corresponding panoramic reprojection $I_t^{\mathrm{gt}}$ and obtain a
reference view $I_t^{\mathrm{ref}}$ from a nearby retained panorama. DiFix additionally receives normalized rendered inverse depth $D_t$
through its confidence branch:
\begin{equation}
\begin{aligned}
\hat{I}_t
&=
F_{\theta}^{\mathrm{DiFix}}
\left(
I_t^{\mathrm{render}},
I_t^{\mathrm{ref}},
\operatorname{Concat}
\left(I_t^{\mathrm{render}}, D_t\right)
\right), \\
\mathcal{L}_{\mathrm{DiFix}}
&=
\left\|\hat{I}_t-I_t^{\mathrm{gt}}\right\|_1
+
\operatorname{LPIPS}
\left(\hat{I}_t,I_t^{\mathrm{gt}}\right).
\end{aligned}
\end{equation}

\noindent\textbf{Temporally consistent video restoration.}
As illustrated in Fig.~3 (left), we reproject each held-out panorama into perspective frames along a rotational trajectory with a fixed camera center. These frames provide ground-truth RGB supervision for video prior adaptation. Rendering the 3DGS at the same camera
poses yields aligned, artifact-prone RGB and depth sequences. We construct training clips of $T$ frames and condition Wan-VACE on the rendered RGB-D sequence and $K$ real reference views selected from retained panoramas. Let $E(\cdot)$ denote the frozen VAE encoder.
The reference views are encoded into a latent conditioning prefix $Z_{1:K}^{\mathrm{ref}}$, and the geometric control is constructed as
\begin{equation}
Z^{\mathrm{ctrl}}
=
\operatorname{Concat}
\left(
E\left(I_{1:T}^{\mathrm{guide}}\right),
E\left(\operatorname{Repeat}_3(D_{1:T})\right),
\tilde{M}_{1:T}
\right),
\end{equation}
where $\tilde{M}_{1:T}$ encodes depth validity.
Given the ground-truth latent
$Z_0=E(I_{1:T}^{\mathrm{gt}})$ and noise
$\epsilon\sim\mathcal{N}(0,I)$, we construct the noisy input
$Z_{\sigma}=(1-\sigma)Z_0+\sigma\epsilon$.
Wan-VACE predicts the target flow $\epsilon-Z_0$ under the RGB-D control and reference conditioning:
\begin{equation}
\mathcal{L}_{\mathrm{Wan}}
=
\mathbb{E}_{Z_0,\epsilon,\sigma}
\left[
\left\|
F_{\psi}^{\mathrm{Wan}}
\left(
Z_{\sigma},
\sigma,
Z^{\mathrm{ctrl}},
Z_{1:K}^{\mathrm{ref}}
\right)
-
(\epsilon-Z_0)
\right\|_2^2
\right].
\end{equation}

During adaptation in Fig.~3 (left), $I^{\mathrm{guide}}$ is the original
3DGS rendering. During trajectory densification in Fig.~2(c), it is replaced by the DiFix-restored RGB sequence, while the rendered depth is retained as geometric conditioning.

\subsection{Generative Trajectory Densification}
\label{sec:trajectory_densification}

After in-the-wild adaptation, the image and video generative models acquire strong priors for restoring residential indoor scenes. Starting from the 3DGS $\Theta_0$ optimized by Correspondence-Guided Geometric Optimization, we construct a continuous camera trajectory $\{P_t\}_{t=1}^{T}$ between each pair of nearby co-visible panoramas. We then perform generative trajectory densification in two stages, as illustrated in Fig.~2(c).

During iterations 7K--15K, we render intermediate views from $\Theta_0$ and restore them using DiFix, conditioned on the rendered depth and reference views. The restored RGB observations are distilled into the 3DGS, producing $\Theta_1$. During iterations 15K--30K, we render RGB-D observations from $\Theta_1$ and apply the full image--video restoration pipeline:
\begin{equation}
\begin{aligned}
    (I_t^{r}, D_t^{r}) &= \mathcal{R}(\Theta_1, P_t), \\
    I_t^{c} &= F_{\mathrm{DiFix}}
    (I_t^{r}, D_t^{r}; I^{\mathrm{ref}}), \\
    \tilde{I}_{1:T} &= F_{\mathrm{Wan}}
    (I_{1:T}^{c}, D_{1:T}^{r}; I^{\mathrm{ref}}),
\end{aligned}
\end{equation}
where DiFix first suppresses local floaters and rendering artifacts, after which Wan-VACE jointly refines the sequence to improve appearance quality and temporal consistency.

Wan-VACE is adapted using fixed-center rotational clips and applied in the second stage to trajectories interpolated between distinct panorama locations, which can involve both rotation and translation. The DiFix-restored RGB sequences and rendered depth provide frame-aligned conditioning along these trajectories, while the video prior restores appearance and temporal coherence. This transfer does not assume that the model has learned translational camera motion from the rotational clips: the initial 3DGS supplies a shared spatial frame, and its per-frame RGB-D renderings provide view-aligned geometry and appearance cues along the new path. Wan-VACE is therefore asked to restore the conditioned sequence, rather than infer camera motion without guidance. We observe this generalization empirically in Fig.~\ref{fig:generated_views}; it is not a guarantee for trajectories outside the reconstructed region. More visualizations can be found in the Appendix~\ref{sec:trajectory_densification}

For each stage, let $\mathcal{V}$ denote the captured views and $\tilde{\mathcal{V}}$ denote the restored RGB pseudo-observations produced by that stage's generative pipeline. During initial optimization, sparse panorama overlap makes unconstrained patch matching unstable, so the precomputed LRM correspondences anchor geometric supervision. After densification, $\Theta_0$ provides coarse global alignment while intermediate views shorten the baselines and introduce substantially more overlapping image content. Local multi-view comparisons can then be drawn from these overlaps rather than only from patches centered at the original anchors. We therefore release the anchor-only restriction and allow both captured and generated views to participate in multi-view optimization:
\begin{equation}
\Theta^{*}
=
\arg\min_{\Theta}
\mathcal{L}_{\mathrm{obs}}(\Theta; \mathcal{V})
+
\lambda_{\mathrm{gen}}
\mathcal{L}_{\mathrm{rgb}}(\Theta; \tilde{\mathcal{V}})
+
\lambda_{\mathrm{mv}}
\mathcal{L}_{\mathrm{mv}}
(\Theta; \mathcal{V} \cup \tilde{\mathcal{V}}).
\end{equation}
In the first stage, this optimization is initialized from $\Theta_0$ and yields $\Theta_1$ using DiFix-restored observations. In the second stage, it is initialized from $\Theta_1$ and yields the final representation $\Theta_2$ using Wan-VACE-refined observations.

Unlike the initial reconstruction stage, multi-view supervision is no longer restricted to patches centered at the precomputed correspondence anchors. Instead, the denser view coverage enables geometric consistency to be optimized across overlapping observations. The captured panoramas remain the source of observed-scene fidelity in $\mathcal{L}_{\mathrm{obs}}$, while the generated RGB frames provide weighted pseudo-view supervision. Rendered depth conditions restoration but is not treated as pseudo-ground-truth geometry, avoiding the direct reinforcement of errors in the initial depth.

\subsection{Automated Spatial Data Curation}
\label{sec:data_curation}

Given the densely rendered RGB-D trajectories, we automatically transform each reconstructed environment into spatially grounded training data. We first sample representative views throughout the scene and prompt Qwen3.5-VL-27B to identify visible, spatially groundable object categories. Its predictions are canonicalized and merged into a scene-specific label vocabulary, avoiding the limited coverage of a predefined indoor taxonomy. Conditioned on this scene-specific vocabulary, Boxer~\citep{detone2026boxer} processes posed RGB-D observations to propose globally aligned 3D object instances and bounding boxes.

We subsequently apply geometric and semantic verification to remove unreliable predictions. Geometric filtering rejects boxes with insufficient multi-view support, implausible dimensions, or inconsistent projections, while a second VLM pass examines their multi-view crops to eliminate incorrect categories and duplicate instances. The remaining annotations form a verified instance set containing canonical labels, captions, and world-coordinate 3D bounding boxes.

Finally, following the question-answer templates of Holi-Spatial~\citep{gao2026holi}, we combine appearance descriptions generated by the VLM from multi-view object crops with geometric attributes and spatial relations computed from the verified 3D bounding boxes. These complementary annotations are used to instantiate large-scale 3D grounding samples and spatial question-answer pairs. This automated label--ground--verify pipeline scales semantic annotation to new environments without human-in-the-loop labeling.

\section{Experiments}

\begin{figure*}[t]
    \centering
    \setlength{\abovecaptionskip}{3pt}
    \setlength{\belowcaptionskip}{0pt}
    \includegraphics[width=\textwidth,trim=0 6mm 0 0,clip]{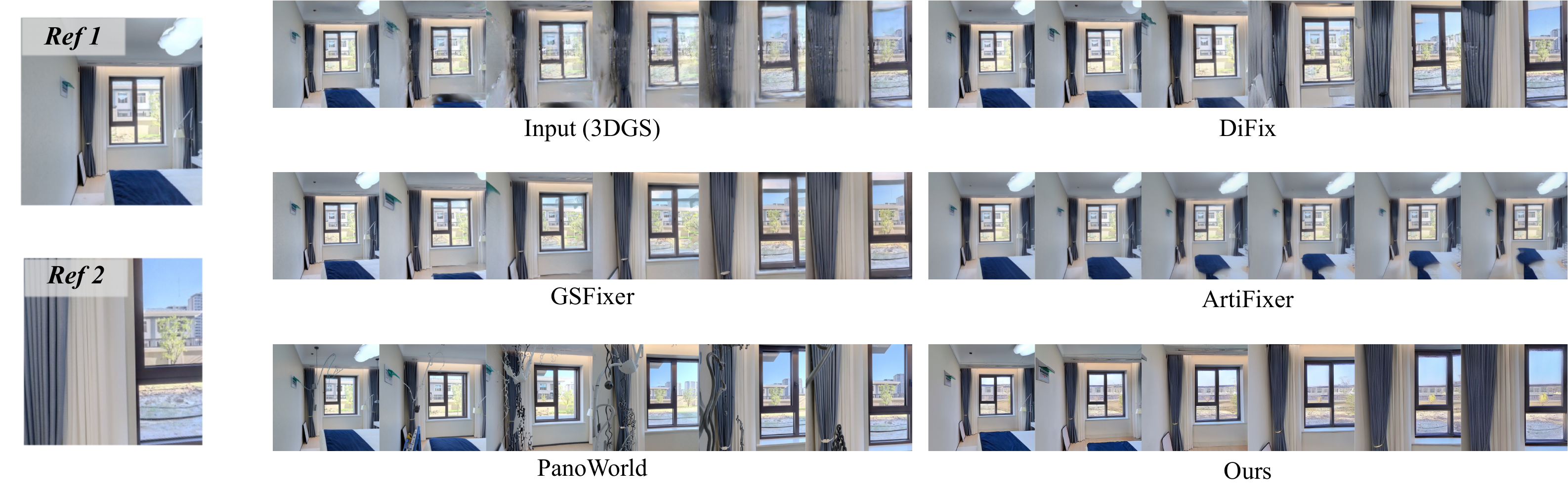}
    \par
    \caption{\textbf{Interpolated-view generation from trajectory endpoints.}
      Ours better preserves scene identity and temporal consistency than existing reconstruction and restoration methods.}
    \label{fig:generated_views}
\end{figure*}

\begin{table*}[t]
\centering
\small
\caption{Novel-view synthesis on three room-random batches. $\uparrow/\downarrow$ indicate better. $^*$ denotes models fine-tuned on our data.
}
\label{tab:room_random_batches}

\setlength{\tabcolsep}{4pt}
\renewcommand{\arraystretch}{1.1}

\resizebox{\textwidth}{!}{%
\begin{tabular}{l|ccc|ccc|ccc}
\toprule[1.2pt]
& \multicolumn{3}{c|}{Room-Random Batch 1}
& \multicolumn{3}{c|}{Room-Random Batch 2}
& \multicolumn{3}{c}{Room-Random Batch 3} \\
Methods
& PSNR$\uparrow$ & SSIM$\uparrow$ & LPIPS$\downarrow$
& PSNR$\uparrow$ & SSIM$\uparrow$ & LPIPS$\downarrow$
& PSNR$\uparrow$ & SSIM$\uparrow$ & LPIPS$\downarrow$ \\
\midrule

\multicolumn{10}{l}{\textbf{Reconstruction Baselines}} \\
\midrule

3DGS
& 19.68 & 0.797 & 0.365
& 18.63 & 0.738 & 0.410
& 19.54 & 0.777 & 0.407 \\

2DGS
& 16.99 & 0.752 & 0.393
& 17.47 & 0.725 & 0.420
& 17.93 & 0.768 & 0.417 \\

PGSR
& 17.32 & 0.726 & 0.439
& 16.71 & 0.646 & 0.483
& 17.51 & 0.684 & 0.479 \\

FSGS
& 20.01 & 0.820 & 0.345
& 18.83 & 0.756 & 0.391
& 19.57 & 0.801 & 0.397 \\

DNGaussian
& 16.73 & 0.769 & 0.423
& 16.38 & 0.725 & 0.460
& 16.16 & 0.767 & 0.464 \\

360-GS
& 18.49 & 0.771 & 0.395
& 17.59 & 0.693 & 0.440
& 18.80 & 0.754 & 0.429 \\

DropGaussian
& 19.56 & 0.820 & 0.345
& 18.55 & 0.761 & 0.387
& 19.52 & 0.804 & 0.391 \\

CoR-GS
& 20.24 & 0.830 & 0.329
& 19.03 & 0.775 & 0.375
& 19.60 & \underline{0.812} & 0.386 \\

Holi-360-recon$^{*}$
& 20.34 & \underline{0.834} & 0.323
& 19.15 & 0.758 & 0.373
& 19.94 & 0.805 & 0.376 \\

\midrule
\multicolumn{10}{l}{\textbf{w/ Diffusion Model}} \\
\midrule

ArtiFixer
& 19.58 & 0.812 & 0.285
& 18.22 & 0.753 & 0.328
& 18.96 & 0.798 & 0.330 \\

PanoWorld$^{*}$
& 18.34 & 0.737 & 0.430
& 17.59 & 0.681 & 0.455
& 18.05 & 0.725 & 0.459 \\

GSFixer
& 18.89 & 0.790 & 0.388
& 17.98 & 0.737 & 0.412
& 18.91 & 0.766 & 0.425 \\

ArtifactWorld
& 19.14 & 0.796 & 0.389
& 17.74 & 0.731 & 0.434
& 17.80 & 0.763 & 0.445 \\

Difix(3DGS)
& 19.55 & 0.768 & 0.337
& 18.27 & 0.709 & 0.362
& 19.09 & 0.754 & 0.349 \\

Holi-360$^{*}$
& \textbf{22.58} & \textbf{0.844} & \textbf{0.231}
& \textbf{21.67} & \textbf{0.785} & \textbf{0.273}
& \textbf{22.21} & \textbf{0.816} & \textbf{0.295} \\

\bottomrule[1.2pt]
\end{tabular}%
}
\end{table*}

\begin{figure*}[t]
    \centering
    \includegraphics[width=\textwidth]{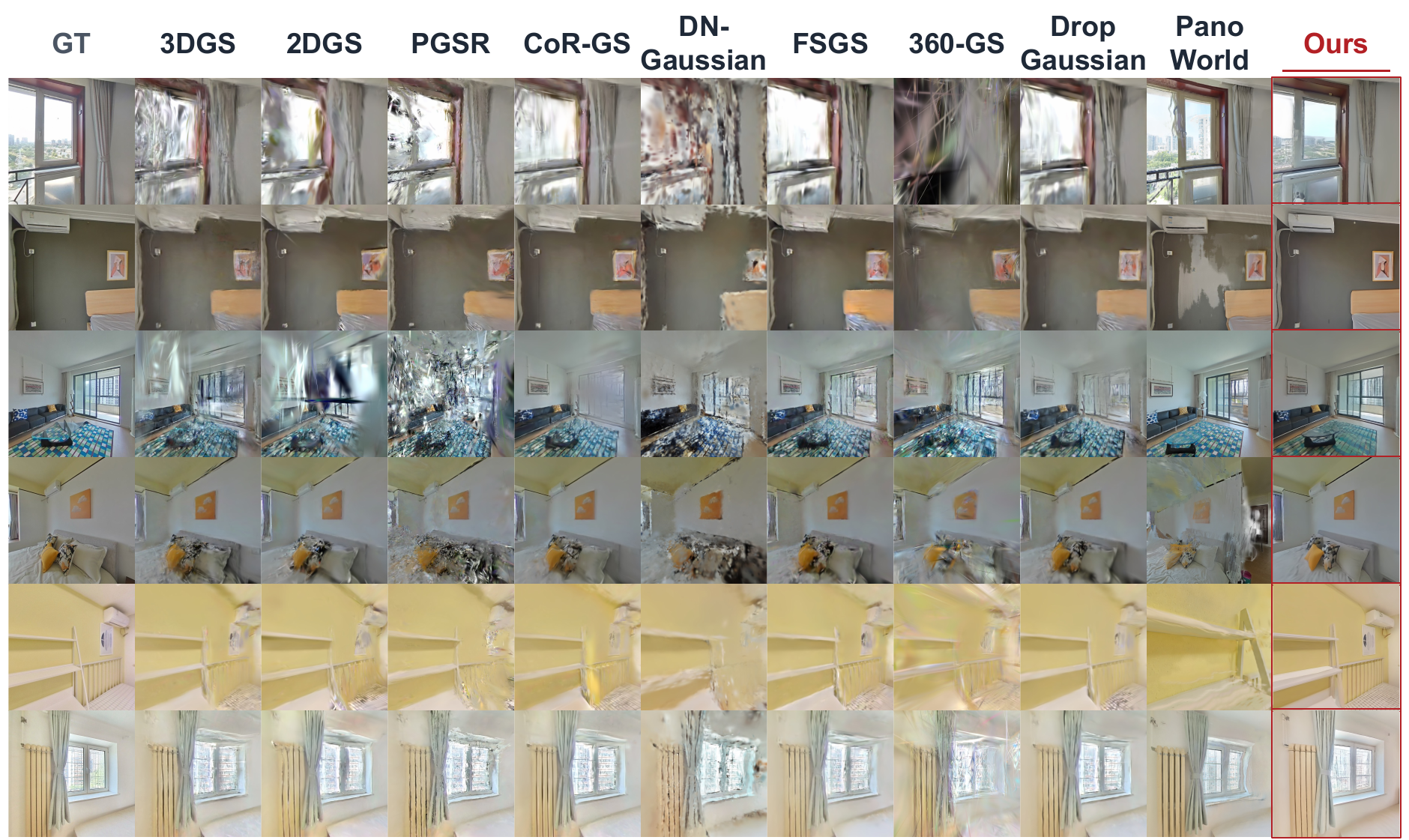}
    \caption{\textbf{Gaussian-rendered out-of-distribution novel views.}
    Columns compare methods; rows share target viewpoints.}
    \label{fig:gaussian_rendering_comparison}
\end{figure*}

\subsection{Out-of-Distribution Rendering Quality}

Table~\ref{tab:room_random_batches} shows that our method outperforms reconstruction-only and generative baselines on held-out household views. Fig.~\ref{fig:gaussian_rendering_comparison} shows that anchored 3DGS better preserves observed structure; PanoWorld, for example, misaligns generated content with the reference in the fourth row. Along interpolated trajectories (Fig.~\ref{fig:generated_views}), DiFix lacks temporal consistency, ArtiFixer propagates opacity artifacts, GSFixer blurs bed boundaries, and PanoWorld introduces reference-misaligned structures. Our geometry-anchored restoration maintains scene identity across views.

\subsection{Geometric Reconstruction}

\begin{figure*}[t]
    \centering
    \includegraphics[
        width=\textwidth
    ]{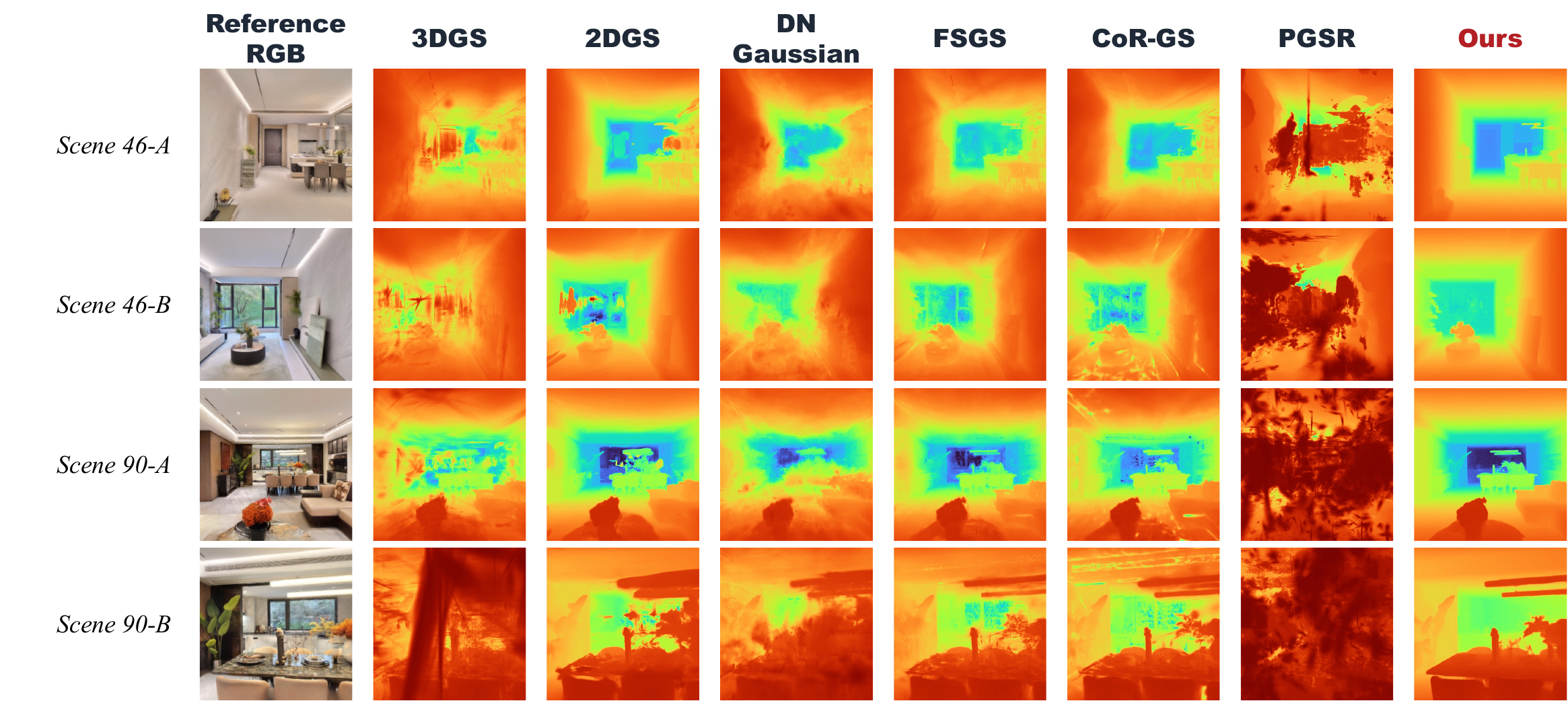}
    \caption{
        \textbf{OOD novel-view depth} over 0.2--10.0\,m.
        Ours produces more complete and coherent geometry than 3DGS, 2DGS, and PGSR.}
    \label{fig:ood_novel_view_depth}
\end{figure*}

We evaluate geometry against LiDAR point clouds from 10 public RealSee3D scenes~\citep{Li2025realsee3d_data}, using the same DA3~\citep{lin2025depth} initialization for all methods. In Tab.~\ref{tab:comparison-2cm}, ours attains the highest F-score at 2\,cm (0.2617 versus 0.2207 for DA3). Under sparse views, RGB-dominated optimization can instead degrade initial geometry when correspondences are unreliable. Figure~\ref{fig:ood_novel_view_depth} confirms that our novel-view depth is more complete and coherent.

\subsection{3D Grounding and VLM Fine-Tuning}
\begin{figure*}[t]
    \centering
    \includegraphics[width=\textwidth]
    {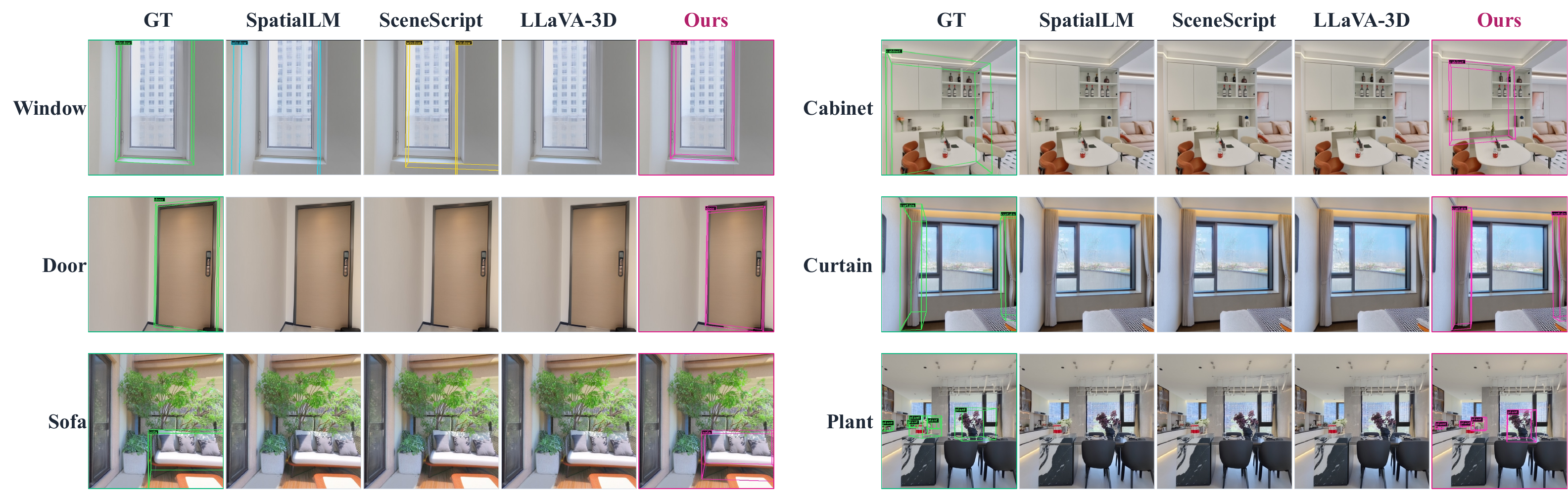}
    \caption{\textbf{3D grounding on real-world scenes.} Qualitative comparison against pseudo ground truth, SpatialLM~\citep{mao2025spatiallm}, SceneScript~\citep{avetisyan2024scenescript}, and LLaVA-3D~\citep{zhu2024llava}. Our verified boxes align more closely with the target objects across diverse categories.}
    \label{fig:grounding_visualization}
\end{figure*}

Tab. ~\ref{tab:grounding_quality} shows that ours reaches 20.66\% Macro F1@25 in 3D grounding, versus 10.40\% for SpatialLM; Fig.~\ref{fig:grounding_visualization} illustrates the box alignment.
We use the 2D segment and lidar point cloud from Realsee to construct the GT grounding; all the implementation details and analysis can be found in Appendix \ref{sec:implement_details}. 
On the Holi360 benchmark (Tab.~\ref{tab:spatial_reasoning_results}),
Qwen3.8-27B~\citep{qwen38} is the strongest zero-shot model, while
fine-tuning Qwen3.5-4B~\citep{qwen3.5} on Holi360 raises classification
macro accuracy from 46.77\% to 85.25\%. Zero-shot models struggle
particularly with metric-scale reasoning: their numerical Acc@0.5m
ranges from 28.95\% to 49.46\%. By providing scale-aware supervision
from reconstructed real-world indoor scenes, Holi360 raises
Qwen3.5-4B's numerical Acc@0.5m from 37.59\% to 76.34\% and reduces
its MAE from 0.9767\,m to 0.3898\,m. Moreover, more details on our QA can be seen in Appendix~\ref{sec:visual_and_QA}.

\begin{table}[t]
\centering

% ==================== Table 3: Left ====================
\begin{minipage}[t]{0.48\textwidth}
\vspace{0pt}
\centering
\caption{\textbf{Geometry reconstruction} at the 2\,cm threshold tested on 10 Realsee scene.}
\label{tab:comparison-2cm}

\resizebox{\linewidth}{!}{%
\begin{tabular}{lccc}
\toprule
\textbf{Method}
& \textbf{P@2cm}
& \textbf{R@2cm}
& \textbf{F@2cm} \\
\midrule
DA3
& 0.2669 & 0.1929 & 0.2207 \\
RaDe-GS
& 0.1881 & 0.1197 & 0.1442 \\
2DGS
& 0.1443 & 0.0858 & 0.1063 \\
3DGS
& 0.0960 & 0.0480 & 0.0635 \\
PGSR
& 0.1399 & 0.0410 & 0.0616 \\
\textbf{Ours}
& \textbf{0.3235}
& \textbf{0.2249}
& \textbf{0.2617} \\
\bottomrule
\end{tabular}%
}
\end{minipage}
\hfill
% ==================== Table 4: Right ====================
\begin{minipage}[t]{0.48\textwidth}
\vspace{0pt}
\centering
\caption{\textbf{3D grounding performance on realsee datasets} (\%).}
\label{tab:grounding_quality}

\resizebox{\linewidth}{!}{%
\begin{tabular}{lcccc}
\toprule
\multirow{2}{*}{\textbf{Method}}
& \multicolumn{2}{c}{\textbf{Macro F1}}
& \multicolumn{2}{c}{\textbf{Micro F1}} \\
\cmidrule(lr){2-3}
\cmidrule(lr){4-5}
& \textbf{@25}
& \textbf{@50}
& \textbf{@25}
& \textbf{@50} \\
\midrule
SpatialLM
& 10.40 & 3.48 & 8.19 & 2.26 \\
SceneScript
& 0.04 & 0.00 & 0.21 & 0.00 \\
LLaVA-3D
& 0.00 & 0.00 & 0.00 & 0.00 \\
\midrule
\textbf{Ours}
& \textbf{20.66}
& \textbf{4.81}
& \textbf{18.59}
& \textbf{4.67} \\
\bottomrule
\end{tabular}%
}
\end{minipage}

\vspace{-0.15in}\end{table}

\begin{table}[!t]
\centering
\caption{\textbf{Spatial reasoning on the Holi360 benchmark.}}
\label{tab:spatial_reasoning_results}
\small
\setlength{\tabcolsep}{4pt}
\renewcommand{\arraystretch}{1.12}
\begin{tabular}{@{}lccccc@{}}
\toprule
\textbf{Model}
& \makecell{\textbf{Classification}\\\textbf{Macro} $\uparrow$}
& \makecell{\textbf{Numerical}\\\textbf{Acc@0.5m} $\uparrow$}
& \makecell{\textbf{Numerical}\\\textbf{MAE} $\downarrow$}
& \makecell{\textbf{13-Class}\\\textbf{Macro} $\uparrow$}
& \makecell{\textbf{Scene}\\\textbf{Macro} $\uparrow$} \\
\midrule

VST-SFT-3B           & 48.04 & 30.65 & 1.2444 & 44.02 & 37.90 \\
InternVL3-8B        & 43.10 & 28.95 & 1.2257 & 39.83 & 34.54 \\
Qwen3.8-27B            & 53.15 & 49.46 & 0.7389 & 52.30 & 49.51 \\
Qwen3.5-4B        & 46.77 & 37.59 & 0.9767 & 44.65 & 42.16 \\
\textbf{Qwen3.5-4B+Ours}
                       & \textbf{85.25} & \textbf{76.34} & \textbf{0.3898}
                       & \textbf{83.19} & \textbf{81.27} \\
\bottomrule
\end{tabular}
\end{table}

\section{Ablations}

\begin{table}[!t]
\centering
\begin{minipage}[t]{0.48\textwidth}
\vspace{0pt}
\centering
\caption{\textbf{Rendering and geometry ablation on 5 scenes of Realsee.}
}
\label{tab:rendering_ablation}
\small
\setlength{\tabcolsep}{4pt}
\renewcommand{\arraystretch}{1.12}
\begin{tabular}{@{}ccccc@{}}
\toprule
ID & \makecell{Gen.\\views} & \makecell{Corr.\\guide}
& PSNR$\uparrow$ & F@2cm$\uparrow$ \\
\midrule
1 & \xmark & \xmark & 19.2374 & 0.1991 \\
2 & \xmark & \cmark & 19.6408 & 0.3297 \\
3 & \cmark & \cmark & \textbf{21.3035} & \textbf{0.3529} \\
\bottomrule
\end{tabular}
\end{minipage}\hfill
\begin{minipage}[t]{0.48\textwidth}
\vspace{0pt}
\centering
\caption{\textbf{Grounding curation ablation on 5 scenes of Realsee.}}

\label{tab:verifier_merge_ablation}
\small
\setlength{\tabcolsep}{5pt}
\renewcommand{\arraystretch}{1.12}
\begin{tabular}{@{}cccc@{}}
\toprule
Verifier & Merge & F1@0.25 & F1@0.50 \\
\midrule
\xmark & \xmark & 16.81 & 8.40 \\
\xmark & \cmark & 18.14 & 9.05 \\
\cmark & \xmark & 18.02 & 9.01 \\
\cmark & \cmark & \textbf{19.42} & \textbf{9.71} \\
\bottomrule
\end{tabular}
\end{minipage}
\end{table}
\vspace{-0.15in}

\noindent\textbf{Rendering and geometry.} In Tab.~\ref{tab:rendering_ablation}, correspondence guidance raises F@2cm from 0.1991 to 0.3297 by stabilizing sparse-view geometry; generated views further raise PSNR from 19.64 to 21.30 and F@2cm to 0.3529. More details on Visualization can be seen in our Appendix ~\ref{sec:vis_ablation}.

\begin{figure}[H]
    \centering
    \includegraphics[
        width=\linewidth,
        trim=0 4mm 0 0,
        clip
    ]{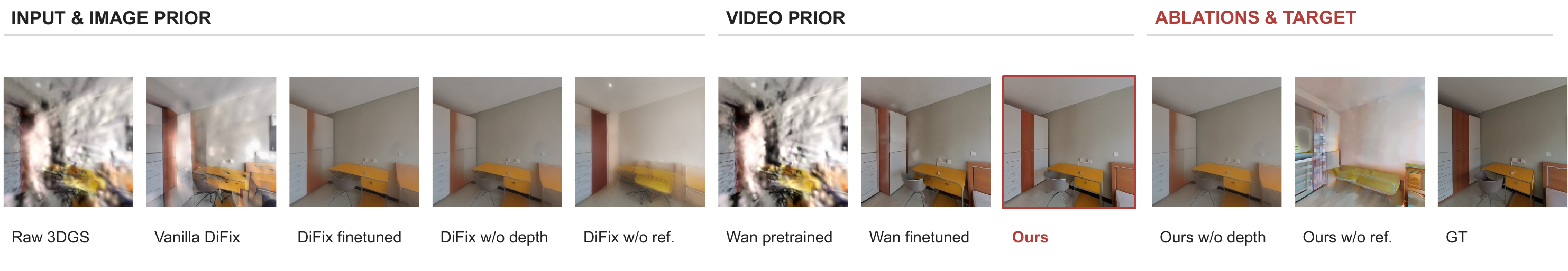}
    \caption{Ablation study on DiFix and Wan; DiFix and Wan finetuned denote DiFix and Wan models after our finetuning. W/O depth denotes only GS RGB input without depth.}
    \label{fig:ablation-difix-wan}
\end{figure}
    \vspace{-4pt}
\noindent\textbf{Generative restoration.} As shown in Fig.~\ref{fig:ablation-difix-wan}, fine-tuning on our in-the-wild data visibly improves both DiFix and Wan-VACE. However, each retains a distinct failure mode: Finetuned DiFix introduces unnatural structures around the left cabinet, while adapted Wan-VACE preserves the overall layout but produces a spurious cabinet protrusion under the influence of input floaters. Combining the two yields substantially cleaner and more structurally coherent views.

\noindent\textbf{Grounding curation.} Both cross-view merging and verification improve grounding in Tab.~\ref{tab:verifier_merge_ablation}, increasing F1@0.25 from 16.81 to 19.42 when combined. More details on Visualization can be seen in our Appendix ~\ref{sec:vis_ablation}.

\section{Conclusion}
In this paper, we presented \textbf{Holi-Spatial 360}, an automated pipeline that turns sparse panoramas of multi-room homes into high-fidelity 3D environments and holistic spatial annotations. Correspondence-guided 3DGS reconstruction provides a coherent geometric foundation, while adapted image and video generative priors restore appearance and complete continuous views without losing the observed scene content. The resulting Holi360 dataset contains over 1K scenes, 100K instance annotations, and 2M spatial question-answer pairs. Experiments show improved reconstruction and annotation quality under sparse panoramic inputs, and demonstrate that Holi360 benefits VLM spatial reasoning.

\subsection*{AI use statement}
We used generative AI tools to improve the language and clarity of this manuscript and to assist with code development. The authors reviewed and revised all AI-assisted text and code, and adapted code suggestions to the specific requirements of our pipeline. The use of generative models and VLMs within the proposed reconstruction and data curation framework is described in the Method section. The authors take full responsibility for the final content, implementation, and reported results.

\section*{Reproducibility Statement}
Section~3 describes our panorama-based reconstruction, generative trajectory densification, and automated spatial data curation pipeline. The Appendix specifies the scene-disjoint data splits, RealSee evaluation annotations, Wan-VACE conditioning, and shared reconstruction initialization. Section~4 reports rendering, geometry, grounding, and spatial-reasoning results, together with component ablations.

\section*{Ethics Statement}
Our source panoramas come from publicly accessible real-estate listings, which may depict private living spaces. Public accessibility does not imply permission to redistribute them. Any release will therefore be limited to materials we are permitted to share, with identifying details such as faces, addresses, and personal documents screened out. Automatically reconstructed geometry and generated annotations may contain errors or biases; they should not be used to identify occupants, surveil homes, or support safety-critical decisions.

\bibliography{iclr2027_conference}
\bibliographystyle{iclr2027_conference}
\clearpage
\appendix
\section{Implementation Details}
\label{sec:implement_details}
\begin{figure}[t]
    \centering
    \includegraphics[width=\linewidth]{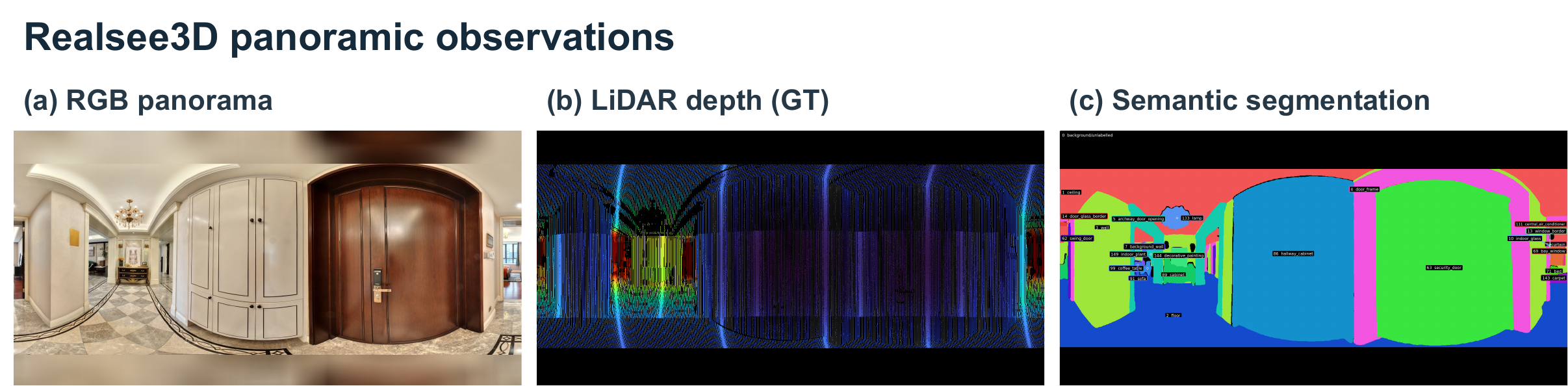}
    \caption{Realsee 2D iamges Sample.}
    \label{fig:sample}
\end{figure}

\begin{figure}[t]
    \centering
    \includegraphics[width=\linewidth]{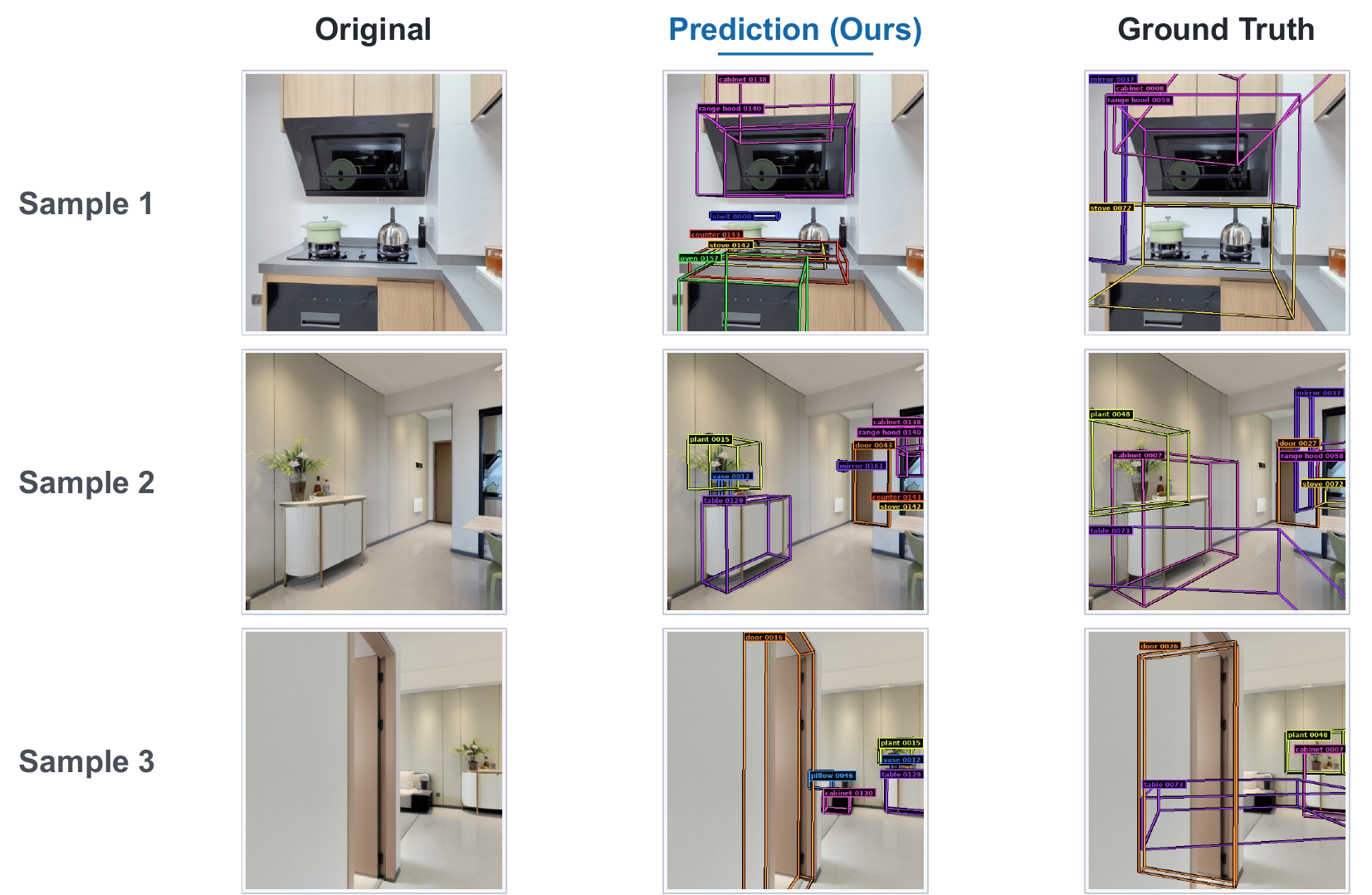}
    \caption{Qualitative comparison between pseudo 3D bounding-box ground truth and our predicted annotations.}
    \label{fig:sample_3d}
\end{figure}

\paragraph{Data collection and splits.}
We ensure that no scene overlaps between the training data and any of the evaluation sets, including RealSee. We collect approximately 1,000 real-world multi-room scenes from publicly available online real-estate and rental platforms across multiple geographic regions. From this collection, we reserve 30 scenes exclusively for testing and exclude them from training. The test scenes are divided into three batches, denoted as Batch 1, Batch 2, and Batch 3, with 10 scenes in each batch. Since these panoramas are captured under real-world conditions, different views may exhibit exposure and illumination inconsistencies. Consequently, a reconstruction may recover the correct scene structure while showing a slight color discrepancy from the reference image, resulting in relatively low PSNR values across all methods.

\paragraph{RealSee evaluation.}
We additionally evaluate on RealSee because, to the best of our knowledge, it is the only publicly available dataset of real-world multi-room indoor scenes that provides both LiDAR measurements and semantic annotations. Nevertheless, RealSee has several limitations. First, its LiDAR observations are relatively sparse and therefore capture only part of the scene geometry. Second, its semantic annotations do not follow a fully unified category taxonomy. For example, labels such as \texttt{wall} and \texttt{background wall}, \textit{door} and \textit{swing door} may refer to closely related semantic categories. We lift the 2D semantic annotations into 3D using the available depth observations and construct the ground-truth 3D bounding boxes used for evaluation. However, sparse depth measurements and inconsistent label granularity can introduce inaccuracies during this projection process.

\paragraph{Reconstruction initialization.}
For a fair comparison, all methods in the reconstruction experiments, including ours, use DA3 as their initialization.

\section{Ablation visualization}
\label{sec:vis_ablation}
\begin{figure*}[t]
    \centering
    \includegraphics[width=\textwidth,height=0.88\textheight,keepaspectratio]{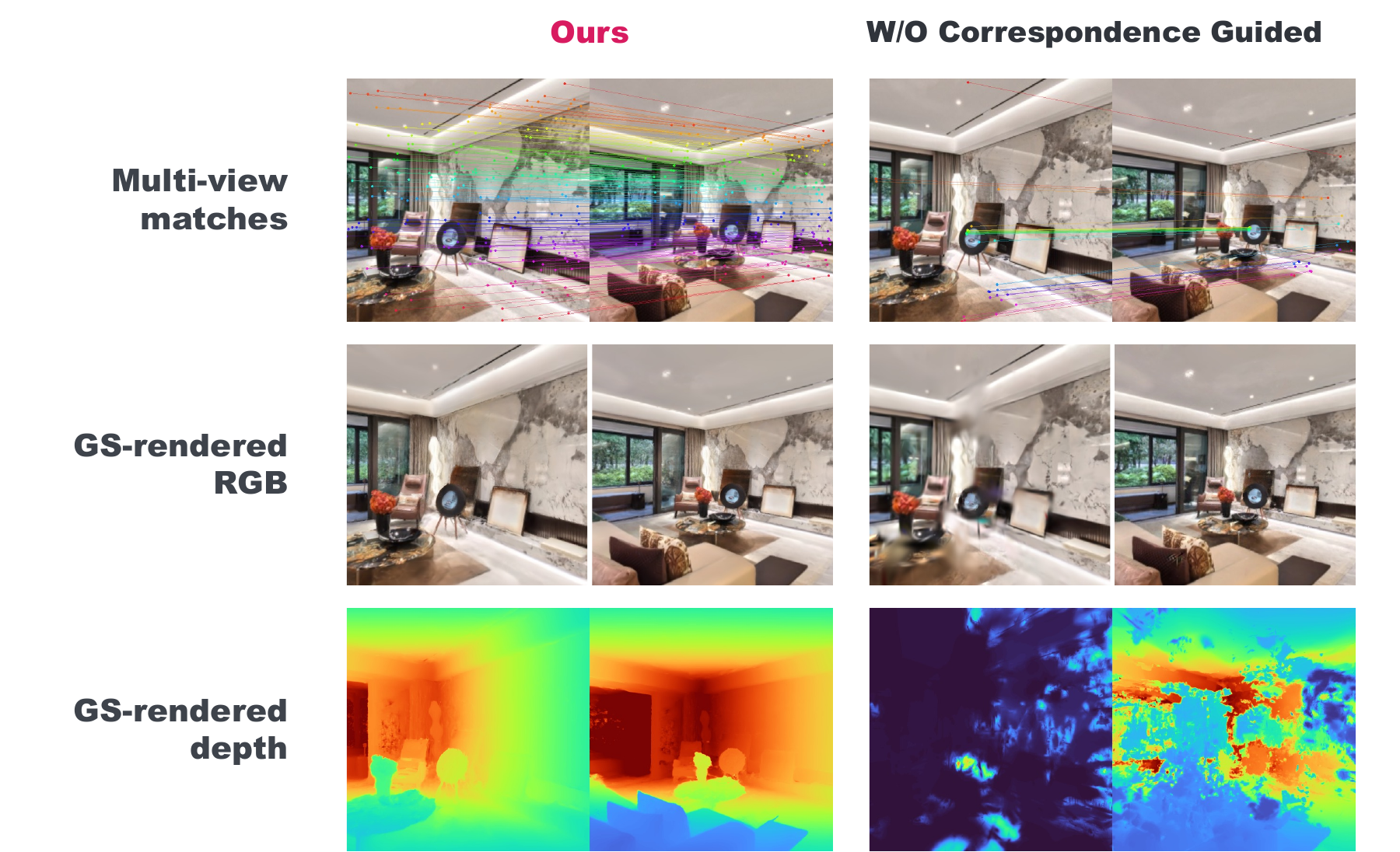}
    \caption{Ablation of the correspondence-guided module in our LRM. All visualized correspondences have reprojection errors below 1 pixel.}
    \label{fig:LRM}\end{figure*}

In Fig~\ref{fig:LRM}, We qualitatively visualize and quantitatively evaluate how the designs of our reconstruction affect the overall pipeline quality. It‘s clear that after the LRM corresononded-guided Gaussian depth can under stable geometry in the sparse view setting, if without it, even under the same initialization, GS optimization becomes dominated by the RGB reconstruction objective, reducing reliable correspondences and weakening multi-view constraints. Consequently, the model overfits the RGB observations and produces more floaters. In detail, as shown in the w/o correspondence-guided case, removing the matching constraint can produce clearly incorrect correspondences even when the reprojection constraint is satisfied. Existing multi-view optimization typically estimates correspondence errors directly between pairs of frames. Once an erroneous match is introduced, it may be repeatedly reinforced during optimization, especially under sparse-view settings where only a few nearby observations are available. For example, the blue point on the panel adjacent to the wall is matched to an incorrect location in the other view, although the resulting correspondence still yields a low reprojection error. This ambiguity motivates our correspondence-guided design, which provides additional constraints to suppress geometrically plausible but semantically incorrect matches. While in Fig~\ref{tab:verifier_merge_ablation}, despite its strong performance, Boxer still produces duplicate detections of the same object and occasional misclassifications. Our post-processing pipeline is designed to address these issues, further improving its accuracy and robustness.

\section{Additional Visualizations and QA templates}
\label{sec:visual_and_QA}
Here in Fig~\ref{fig:visual_supp}, we present a large-scale qualitative visualization of generated views, all sampled from the test set. And we also visualize our QA example in Fig~\ref{fig:qa}, which contains normal RGB understanding and panorama spatial understanding in real-world indoor scenes.

\begin{figure}[H]
    \centering
    \includegraphics[
        width=\linewidth,
        trim=0 0mm 0 0,
        clip
    ]{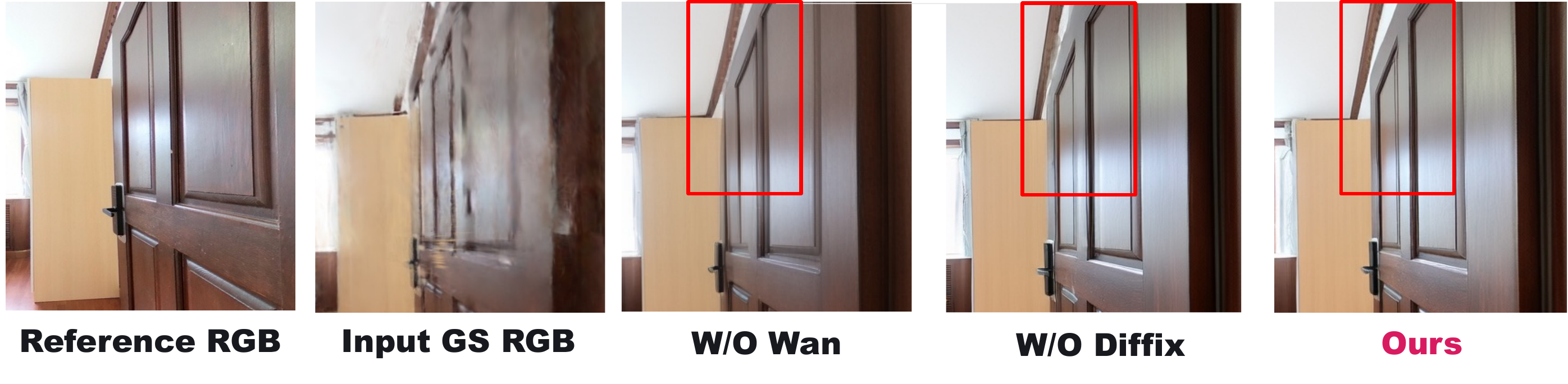}
    \caption{Extra ablation study on DiFix and Wan.}
    \label{fig:ablation-difix-wan-app}
\end{figure}

\begin{figure*}[t]
    \centering
    \includegraphics[width=\textwidth,height=0.88\textheight,keepaspectratio]{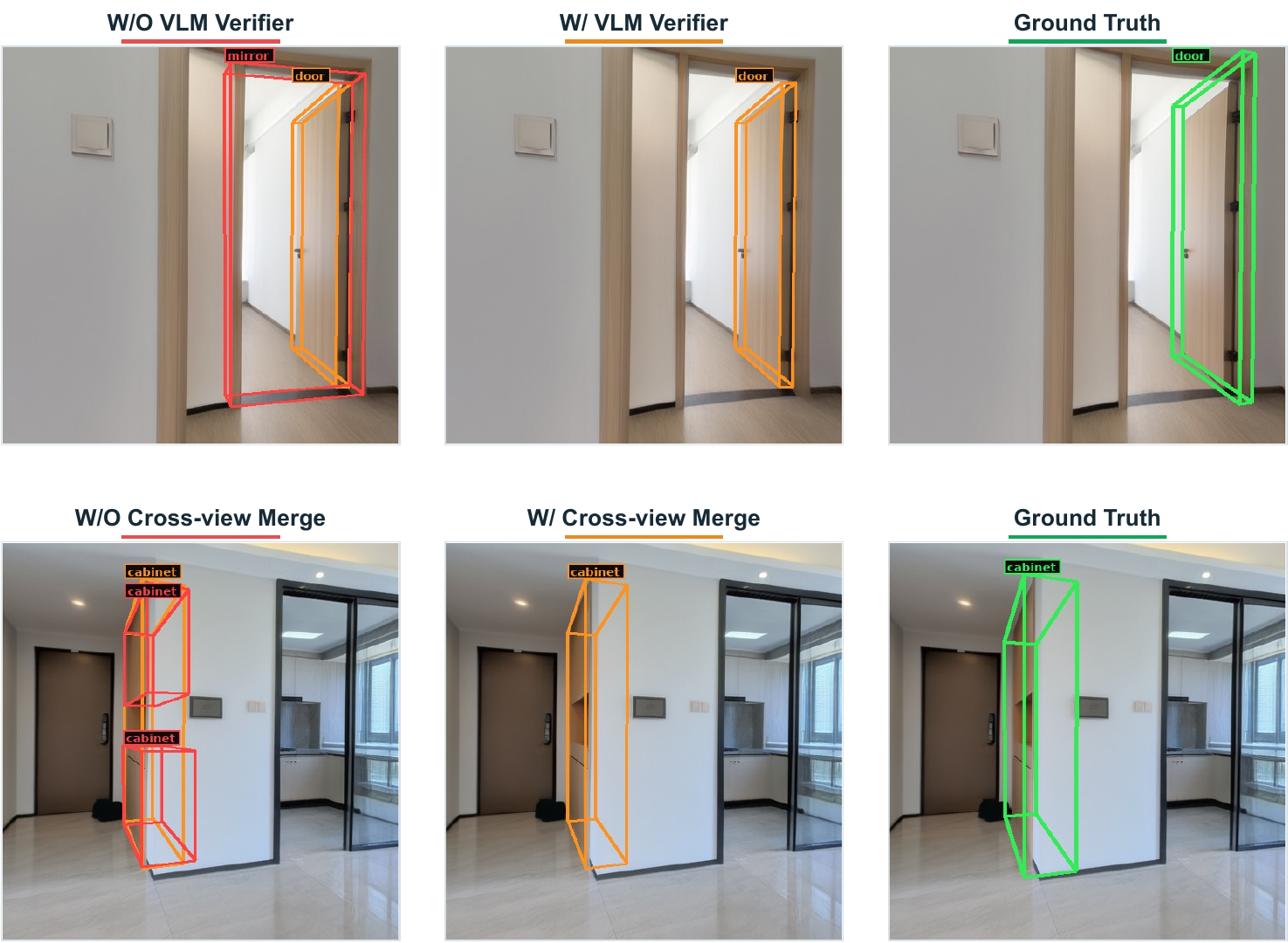}
    \caption{\textbf{Ablation on our cross-view merge and VLM verifier.} }
    \label{fig:ab_data_cura}\end{figure*}

\begin{figure*}[t]
    \centering
    \includegraphics[width=\textwidth,height=0.88\textheight,keepaspectratio]{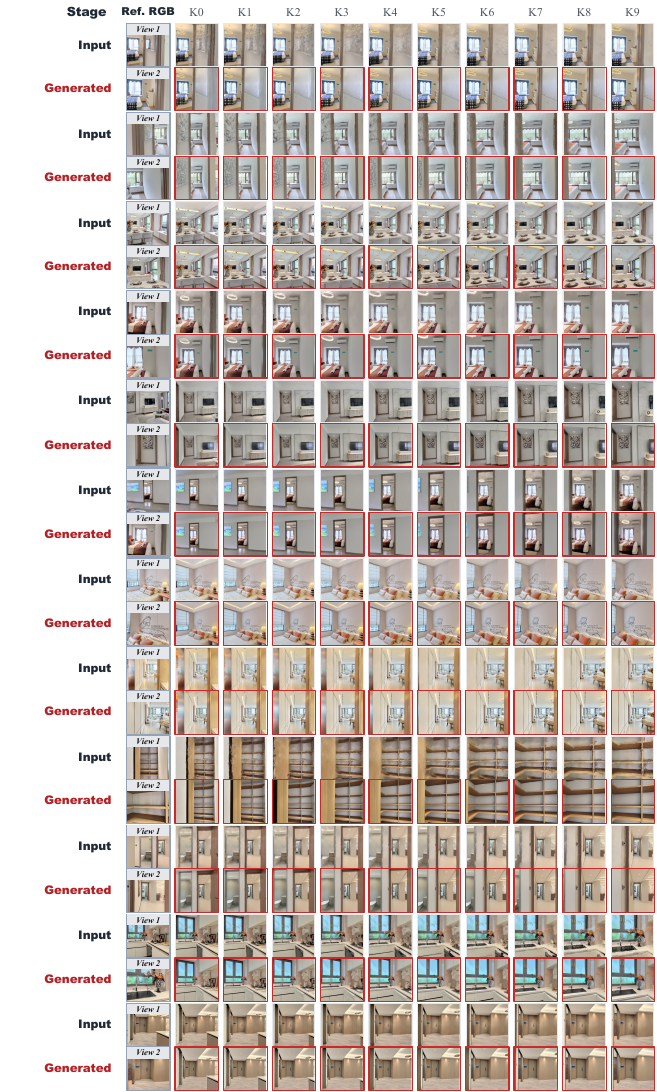}
    \caption{\textbf{Additional qualitative results.} Reference views, 3DGS inputs, and generated trajectories across diverse indoor scenes.}
    \label{fig:visual_supp}\end{figure*}

\begin{table*}[t]
\centering
\caption{Forward--backward-consistent matches ($<1$ px) versus input-view count.}
\label{tab:matching_points}
\begin{tabular}{lccc}
\toprule
Method & 2 Frames & 4 Frames & 8 Frames \\
\midrule
DA3    & \textbf{114.6} & \textbf{87.1} & \textbf{79.0} \\
VGGT   & 46.8 & 22.7 & 15.5 \\
i3dgs  & 44.0 & 44.0 & 44.0 \\
COLMAP & 29.0 & 29.0 & 29.0 \\
\bottomrule
\end{tabular}
\end{table*}

\begin{figure*}[t]
    \centering
    \includegraphics[width=\textwidth,height=0.88\textheight,keepaspectratio]{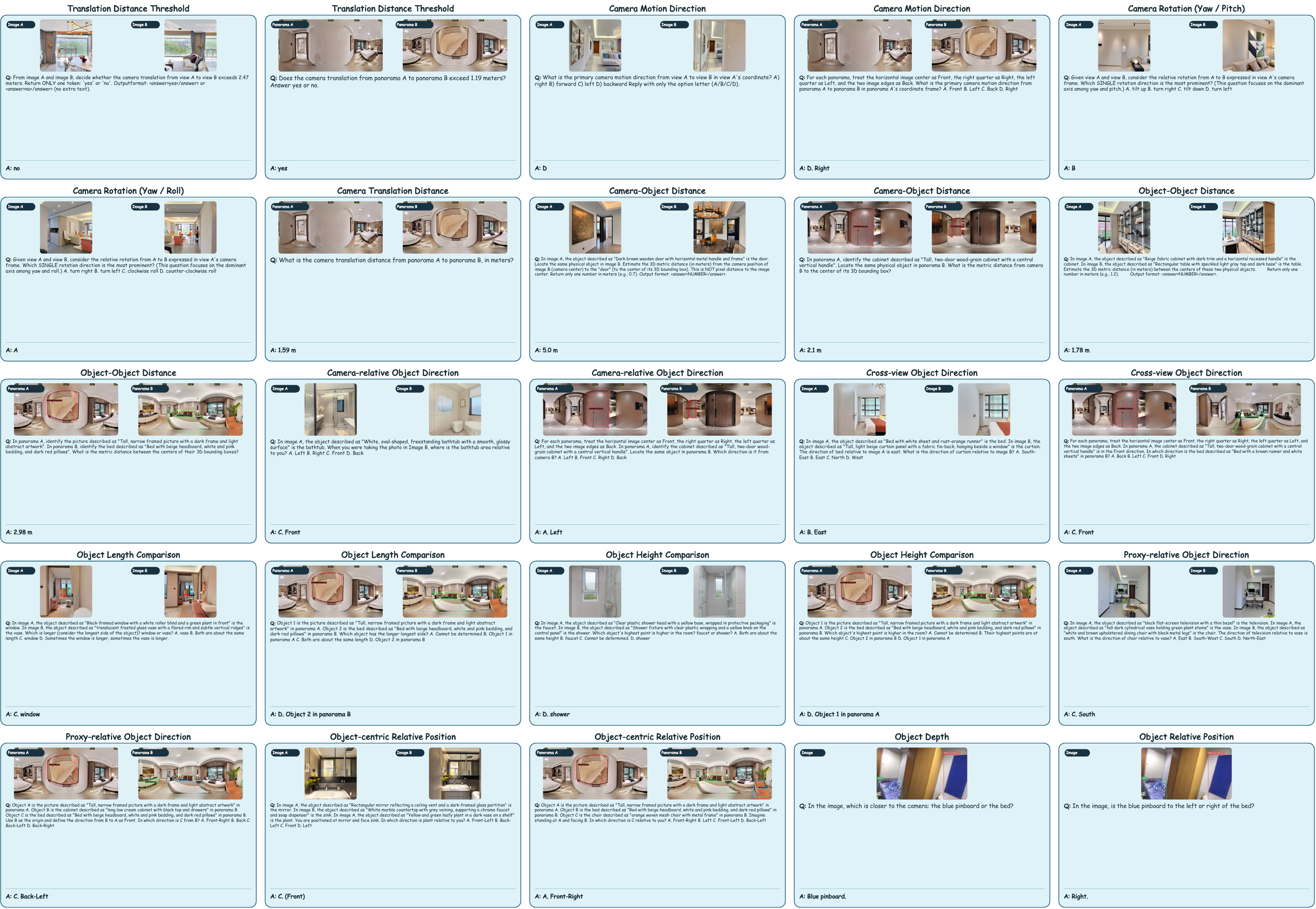}
    \caption{\textbf{Holi360 benchmark QA visualization.}}
    \label{fig:qa}\end{figure*}

\end{document}